\def\year{2022}\relax
\documentclass[letterpaper]{article} 
\usepackage{aaai22}  
\usepackage{times}  
\usepackage{helvet}  
\usepackage{courier}  
\usepackage[hyphens]{url}  
\usepackage{graphicx} 
\usepackage{natbib}  
\usepackage{caption} 
\DeclareCaptionStyle{ruled}{labelfont=normalfont,labelsep=colon,strut=off} 
\usepackage{array, makecell}

\usepackage{algorithm}
\usepackage{algorithmic}

\usepackage{newfloat}
\usepackage{listings}
\floatstyle{ruled}
\newfloat{listing}{tb}{lst}{}
\floatname{listing}{Listing}
\title{ADAPT: An Open-Source sUAS Payload for Real-Time Disaster Prediction and Response with AI}
\author{
    Daniel Davila,\textsuperscript{\rm 1} Joseph VanPelt,\textsuperscript{\rm 1} Alexander Lynch,\textsuperscript{\rm 1} Adam Romlein,\textsuperscript{\rm 1} Peter Webley,\textsuperscript{\rm 2} \\ Matthew S. Brown\textsuperscript{\rm 1}
}
\affiliations{

    \textsuperscript{\rm 1} Kitware Inc., 101 E. Weaver St. Suite G4, Carrboro, NC 27510\\
    \textsuperscript{\rm 2} University of Alaska Fairbanks\\
    adapt-payload@kitware.com
}

\begin{document}

\maketitle

\begin{abstract}

Small unmanned aircraft systems (sUAS) are becoming prominent components of many humanitarian assistance and disaster response (HADR) operations. Pairing sUAS with onboard artificial intelligence (AI) substantially extends their utility in covering larger areas with fewer support personnel. A variety of missions, such as search and rescue, assessing structural damage, and monitoring forest fires, floods, and chemical spills, can be supported simply by deploying the appropriate AI models. However, adoption by resource-constrained groups, such as local municipalities, regulatory agencies, researchers, and indigenous persons, has been hampered by the lack of a cost-effective, readily-accessible baseline platform that can be easily adapted to their unique missions. To fill this gap, we have developed the fully free and open-source ADAPT multi-mission payload for deploying real-time AI and computer vision onboard a sUAS during local and beyond-line-of-site missions. We have emphasized a modular design with low-cost, readily-available components, open-source software, and thorough documentation (\url{https://kitware.github.io/adapt/}). The system integrates an inertial navigation system, high-resolution color camera, computer, and wireless downlink to process imagery and broadcast georegistered analytics back to a ground station. Our goal is to make it easy for the HADR community to build their own copies of the ADAPT payload and leverage the thousands of hours of non-recurring engineering we have devoted to developing and testing this general-purpose capability. In this paper, we detail the development and testing of the ADAPT payload. We also demonstrate the example mission of real-time, in-flight ice segmentation to monitor river ice state and provide more-timely predictions of catastrophic flooding events. We deploy a novel active learning workflow to annotate river ice imagery, train a real-time deep neural network for ice segmentation, and demonstrate operation during field testing.

\end{abstract}

\section{Introduction}
\label{intro}
Over the last decade, economies of scale and changing regulations have substantially reduced the barrier to entry into utilizing drones for a wide variety of operations. This democratization of aerial capabilities, previously reserved for large aircraft and large organizations, is leading to novel opportunities for humanitarian assistance and disaster response. At the same time, advances in deep neural network (DNN) architectures have allowed artificial intelligence (AI) processing to run faster on less-capable hardware, such as readily-available, lightweight, and low-power edge computers. It is now possible to deploy DNN-based computer vision in real time on small unmanned aircraft systems (sUAS)~\cite{zhu2018, Yang2021, Castellano2020_2, Castellano2020}. Such AI-enhanced systems could automatically scan large  or hazardous areas and provide essential situational awareness~\cite{Shihavuddin2019,Kuchhold2018,Singh2018,Jung2018}. In-flight processing capabilities are essential for beyond-line-of-sight operations, where intermittent, low-bandwidth wireless connections preclude streaming of high-resolution raw imagery back to a base station. Instead, compact summary analytics, such as semantic segmentations~\cite{tzelepi2021semantic,Lyu2020} or detected objects~\cite{Kuchhold2018} georegistered on a map, can be streamed to a base station and then beyond to decision makers. These systems are already demonstrating utility in search and rescue operations~\cite{Castellano2020}, including from open water \cite{Lygouras2019}, flooded areas~\cite{Albanese2021}, avalanches~\cite{Bejiga2017}, and dense forest \cite{Yong2018}. Additional applications include monitoring forest fires \cite{Zhao2018,Kinaneva2019,Jiao2019}, live mapping of floods in support of emergency services \cite{s19071486,Munawar2021,su13147547,hashemi2021flood}, assessing structural damage \cite{kang2018autonomous,wu2018coupling,bhowmick2020vision}, monitoring chemical spills \cite{jiao2019new,ghorbani2020identification,de2020oil}, and monitoring reef \cite{ridge2020} and coastal sand dune \cite{choi2017uav} erosion. In all of these applications, low-latency dissemination of results and accurate geolocation is essential to be useful for decision makers.

\begin{figure}[t]
  \centering
  \includegraphics[width=\linewidth]{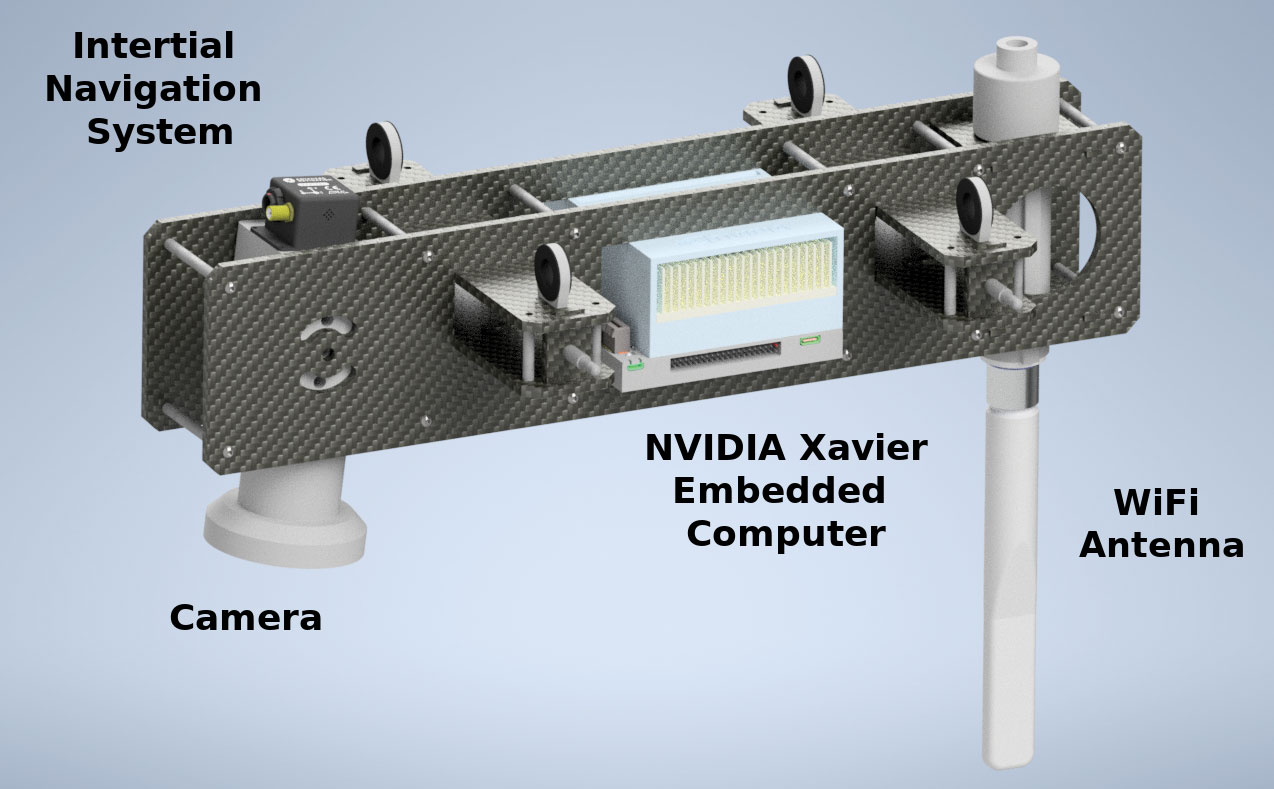}
  \includegraphics[width=\linewidth]{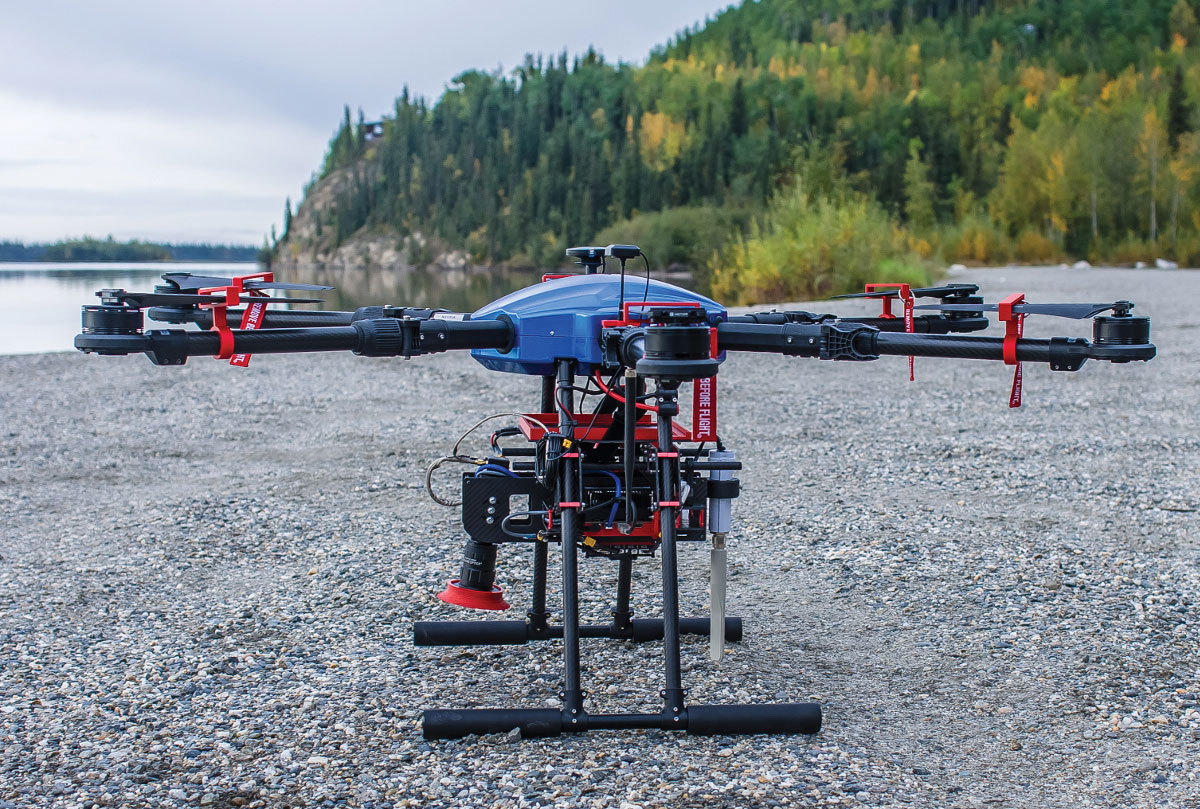}
  \caption{(Top) Rendering of the ADAPT payload assembly CAD model that is publicly available at \url{https://kitware.github.io/adapt/}. (Bottom) Payload deployed on a sUAS.}
  \label{fig:payloadCAD}
\end{figure}

The explosion in readily accessible capabilities of both drones and DNN computer vision has been largely supported by open-source software projects providing tools with free-use licenses. This includes drone components, such as flight controllers (ArduPilot and PX4), communication protocols (MAVLINK), and post-processing tools (OpenDroneMap), as well as open-source projects for general computer vision (OpenCV), machine learning (scikit-learn), deep learning (Pytorch, TensorFlow, Caffe), and data annotation (VIAME, CVAT, LabelMe). These have allowed smaller organizations to build creative solutions on top of these foundational components. Similarly, an important trend within the computer vision research community, which has helped to accelerate progress, is to accompany publications with open-source implementations of their algorithms (e.g., on Github). With this, one has access to all of the tools~\textemdash~and equally important, sufficient and approachable documentation~\textemdash~to build state-of-the-art AI processing pipelines to support humanitarian assistance and disaster response missions. However, there is a lack of commensurate open-source projects, tools, and documentation to support deployment of this processing live, onboard a sUAS with all of the management of imaging and navigation sensors required to produced real-time, georegistered results. This leaves independent groups to redundantly develop mechanical and electrical payload designs and write low-level software for sensor control and data management before they can start to focus on the unique aspects of their missions.

The \textbf{A}utonomous \textbf{D}ata \textbf{A}cquisition and \textbf{P}rocessing \textbf{T}echnologies (ADAPT) Multi-Mission Payload Project\footnote{Project website: \url{https://kitware.github.io/adapt/.}}, which we detail in this paper, was created to provide a foundation of systems integration, open-source software, and thorough documentation for organizations to deploy readily-available AI capabilities onboard a sUAS in a way that allows rapid dissemination of results. As its name implies, payload capabilities can be adapted to specific missions simply by swapping in different DNN models and sensors. For missions without an existing corpus of data and trained models, the ADAPT payload and associated infrastructure can support optimized data collection and annotation to train the requisite models. Then, the same system can be used for aerial, real-time deployment of those models. 

We make this open-source system (both hardware and software) free to the community so that stakeholders may build upon the designs as they identify minor adaptations needed to support a richer set of mission requirements. Our hope is that the community will share these findings and adaptations with one another via contributions to the open-source ADAPT project, thereby reducing the overall burden of non-recurring payload engineering.

While the ADAPT payload is intended as a general-purpose tool, we focus this paper on the specific mission of monitoring the evolution of frozen rivers in Alaska to better understand and predict catastrophic flooding events. When the Alaskan winter snow melt is rapid and river ice is mostly intact, immense pressure builds on the river ice, leading to a dramatic mechanical breakup~\cite{BeltaosSpyros2003}. Mechanical breakup occurred along important waterways in Alaska during the spring of 2013, including one event along the Yukon River that led to the devastating impacts at Galena, Alaska ~\cite{Galena2013,Taylor2016}. High water and ice hit many villages along the river (Eagle, Circle, and Ft. Yukon), but Galena suffered the most. Colossal damage left a heavy economic toll with 169 houses destroyed and an estimated \$6 million needed to rebuild the public infrastructure~\cite{Galena2013}. 

Historically, breakup-flood forecasting relies on manned reconnaissance flights to issue timely flood watches and warnings, such as the National Weather Service's \textit{River Watch Program}. These flights provide observations to track river ice conditions (strength, extent, and movement) over large regions (tens to hundreds of miles) and identify ice jams.  The vast extent of Alaska, with limited ground observations, makes it difficult to provide timely and accurate forecasts. This leads to an over-reliance on lower-resolution, satellite-derived products subject to extended occlusion from clouds. Alternatively, low-cost sUAS deployments of the ADAPT payload could allow organizations and indigenous communities to monitor river conditions, leveraging real-time, in-flight processing to rapidly disseminate results over low-bandwidth wireless links.

The main contributions of this work are: 

\begin{itemize}
  \item A free and open-source sUAS payload for deploying real-time, in-flight AI, enabling various missions supporting humanitarian assistance and disaster response 
  \item A framework achieving accurate temporal synchronization of imagery and GPS/INS streams for exploitation of visio-inertial odometry and environment mapping 
  \item A publicly available trained model for river ice segmentation for use in further research and development 
  \item A novel active learning annotation workflow 
  \item A demonstration of payload utility in collecting quality imagery, validated in flight 
\end{itemize}

\section{Open-Source sUAS Payload}
\label{opensource}
Most demonstrated sUAS payloads are commercial products tightly integrated into the drone itself with limited end-user customization options or are one-off research systems to support a highly specific mission. This leads to many different groups spending large amounts of time and money on development and integration. Our goal in starting the open-source ADAPT project is to provide a basis of integrated camera and inertial navigation sensors required for any organization to get started in AI-enhanced sUAS missions. We provide a baseline implementation of the system with low-cost, readily-available components for collecting RGB imagery, but the overall design allows for great flexibility in reconfiguration, such as exploiting other camera modalities (e.g., infrared, multispectral) and different wireless radios.

\begin{table*}[t!]
\centering
\begin{tabular}{|l|l|c|}
    \hline
    \thead{\textbf{Component}}  & \thead{\textbf{Description}} & \thead{\textbf{Price (USD)}} \\
    \hline
    Advanced Navigation Spatial & Intertial navigation system & \$3,230 \\
    FLIR Blackfly BFS-PGE-161S7C-C & Global shutter color camera & \$1,600 \\
    Edmund Optics 86-569 & Lens & \$995 \\
    NVIDIA Jetson Xavier AGX & Embedded computer with GPU  & \$750 \\
    Frame & Carbon fiber cutouts & \$500  \\
    Ubiquiti Bullet AC & Dual-band radio & \$182  \\
    Samsung MZ-V7E1T0BW & M.2 NVMe SSD (1TB) storage & \$150  \\
    Ubiquiti LiteBeam Bridge & Ground station wireless bridge & \$80  \\
    Misc & Hardware (fasteners, cables, etc.) & \$75  \\
    Edmund Optics 88-065 & Cable (Hirose) for camera & \$67  \\
    StarTech ST1000SPEX2 & PCIe network card & \$27  \\
    TrendNET TEW-AO57 & Antenna n-type dual & \$15  \\
    PoE Texas PoE injector & Power for antenna & \$8  \\
    \hline
    \textbf{Payload Total} &  &  \textbf{\$7,679}  \\
    \hline

\end{tabular}
\caption{Components required to complete payload build. We also integrated a cheaper, rolling-shutter camera, the \$720 Blackfly BFS-PGE-200S6C-C with \$550 Computar V0826-MPZ lens.}
\label{payload-hardware-table}
\end{table*}

Figure~\ref{fig:payloadCAD} shows the major components of the payload: the camera, inertial navigation system (INS), NVIDIA Xavier computer, and WiFi antenna. The full set of components is detailed in Table~\ref{payload-hardware-table}. The system shown can be sourced and assembled for less than \$7,700 (USD) by technicians without expert knowledge of complex systems integration. The frame of the payload is cut from carbon-fiber sheets, and components are affixed to this frame with 3-D printed parts. The camera mount is 3-D printed from Nylon+CF filament for strength and rigidity, as the INS is also screwed into this mount. The camera can be manually rotated before the flight to operate anywhere from nadir to forward-looking, and this modular design easily supports different cameras via modifications to the 3-D printed mount. PLA filament was used to 3-D print holders for the Xavier and WiFi antenna, while TPU was used for the Lens Hood. Fully assembled the payload weighs approximately 2.6 kg and draws approximately 68 W of power while in operation. All CAD models for these components are made publicly available (\url{https://kitware.github.io/adapt/}), allowing them to be affordably manufactured in-house or from various online rapid-prototyping services.

Whenever possible, industry standards (e.g., Genicam) were used in the electrical and software configuration. We chose GigE-interface cameras over USB3 cameras, as USB3 is known to cause substantial electromagnetic interference with the GPS signal. The Xavier edge computer runs Ubuntu 18.04. The software stack utilizes the Robot Operating System (ROS) for flexibility in adding processing nodes, supporting inter-system communications (e.g., between payload and base-station computers), and integrating additional sensors, as many hardware components that one might put on a robot already have ROS drivers~\cite{ROS}. 

\subsection{Time Synchronization, Calibration, and Georegistration}
\label{timesync}

A critical requirement across most envisioned missions is placing AI analytics derived from individual images (e.g., segmentation maps or detected object bounding boxes) within a geospatial context. The INS reports its pose (location and orientation) with respect to the world as a function of time ($\sim$100 Hz). The camera and INS are rigidly mounted together, so camera pose associated with each image can be inferred by interpolating INS pose onto each image exposure time and rotating it into the camera frame with a pre-calibrated \textit{INS-to-camera} transformation. With an accurately calibrated camera model~\cite{Hartley2003,colmap}, raw image analytics (specified in image coordinates) can be transformed into georegistered results. This process is called direct georeferencing. However, for this process to be effective, we need accurate temporal synchronization between the navigation data stream and image timestamps. For example, a sUAS turning at 10\textdegree/s and viewing an area of ground 200 meters away will yield $\sim$35-m georegistration error for a one-second time-sync error (erroneously associating the image with INS pose from 1 seconds in the past or future). This is particularly problematic when producing land cover segmentation maps covering an extended area because image-to-image georegistration errors lead to discontinuities and inconsistencies where image fields of view overlap.

Achieving accurate time synchronization across sensors for this purpose is notoriously difficult, often demanding substantial integration efforts including custom microcontrollers~\cite{VersaVIS,okvis}. Alternatively, our approach to time synchronization exploits recently-available machine vision cameras (e.g., FLIR Blackfly) that support the IEEE 1588 Precision Time Protocol (PTP), which can achieve sub-microsecond synchronization. The Xavier edge computer acts as a PTP master, receiving one pulse per second (1PPS) from the INS, with the 1PPS derived from GPS time with sub 0.1 $\mu$s accuracy. The internal clock on the camera is synchronized via PTP to the Xavier's clock. Therefore, the camera, INS, and Xavier clocks are all accurately tied to GPS time, which would even provide time synchronization between physically separate ADAPT payloads concurrently flying. To validate time synchronization, we pointed the camera at an HDMI monitor connected to the Xavier displaying its system time, and the time visible in the image was highly consistent with the image's time stamp within the expected uncertainty due to monitor latency. Additionally, we triggered the camera using the 1PPS directly and found image timestamps to be consistent with whole-number seconds.

To further validate time synchronization, assess overall accuracy of INS outputs, and to calibrate the camera's intrinsic parameters~\cite{Hartley2003}, we used the open-source structure from motion (SfM) software Colmap~\cite{colmap} to process images from a calibration flight. An optimal calibration flight includes figure eights spanning multiple altitudes over a fixed area of ground. Colmap analyzes correspondences between images collected during the flight and automatically recovers camera intrinsic parameters and the pose associated with each image. The SfM-computed poses are generally highly accurate since they are derived from high-resolution imagery. However, monocular SfM solutions are only defined up to an arbitrary similarity transform. Since the camera and INS are nearly co-located, we solve for the similarity transform that best maps the SfM camera positions into the associated INS-reported positions within a local tangent plane coordinate system\footnote{We define an east, north, up coordinate system centered at the median latitude and longitude over the flight and transform INS pose and, using a best-fit similarity transform, the SfM poses into this Cartesian coordinate system.}. We treat the resulting SfM-derived camera orientations as ground truth. We solve for the optimal fixed INS-to-camera rotation, which when applied to INS-reported orientations best recovers the associated truth camera orientations. This calibration is used in future flights for direct georegistration of image analytics. All code associated with this calibration routine has been made publicly available. We can also interrogate our time-sync accuracy by testing whether an artificial time offset produces better agreement, and we found an offset sufficiently near zero, when considering the temporal precision of the technique.

\begin{figure}[!t]
  \centering
  \includegraphics[width=\linewidth]{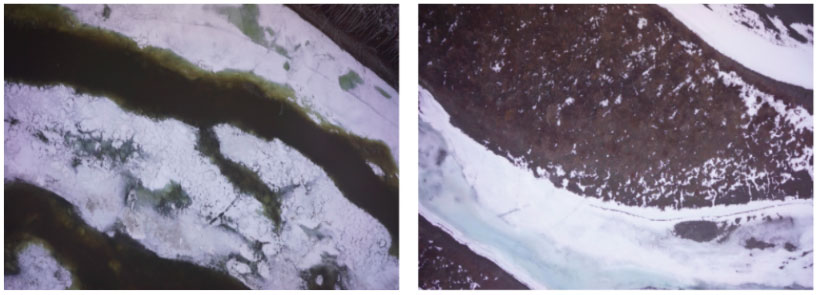}
  \caption{Example imagery collected in the KUAF dataset. The dataset is comprised of over 1000 high resolution images taken from an sUAS deployed over the Yukon River near Circle, Alaska. }
  \label{fig:kuaf}
\end{figure}

For testing the ADAPT payload for a HADR-relevant mission, we investigate the task of river ice monitoring. To our knowledge, the closest representative open dataset in this domain is the river ice dataset from the University of Alberta \cite{alberta_ice}. Other works have compiled similar but unreleased datasets for their work~\cite{ICENET2020}. The Alberta river ice dataset is an important addition to the public space but is very limited in scope, lacking diversity in scale, sufficient examples of river\textendash bank boundaries, and instances of ice buildup\textendash breakup surrounding melt events. To overcome these limitations, we have curated the ANONYMOUS River Ice Dataset (KUAF) as shown in \ref{fig:kuaf}. KUAF consists of over 1000 high-resolution (4000 x 6000 px) RGB color images taken from an aerial drone platform that flew multiple days of the same route over a segment of the Yukon River near Circle, Alaska. The unmanned system was flown at 558 feet above ground level (AGL). The data contains many examples of ice buildup along the river, loose ice, ice\textendash snow and ice\textendash land interfaces, and complex topologies of ice, water, and land within a local area. Additionally, the dataset contains imagery of the same segment of river taken over multiple days of ice accumulation and dissipation. This data can be used in change analysis algorithms. The KUAF River Ice Dataset is under active development and will be released in full to the community at a future time through the ADAPT Github site (\url{https://kitware.github.io/adapt/}).

\subsection{Deep Neural Network Model}
\label{dnn_model}

For the ice segmentation task we base our solution on the recent real-time segmentation work of BiSeNetV2 \cite{bisenetv2}, ideal for the low-SWaP requirements of the ADAPT payload. This work utilizes a novel and compute-efficient dual-stream approach to semantic segmentation. Their network breaks the task into two paths. One is a detail branch, which is channel-rich and responsible for learning dense representations of the low level information. The other is a semantic branch which has shallow layers but quick spatial pooling to aggregate information and context across a wide ROI over the original image. While this is quite applicable to our sUAS payload, it has also been utilized in various other problems, such as general robotic applications \cite{tzelepi2021semantic} and pose measurement \cite{Du2021}, and has inspired similar model architectures for the tasks of road segmentation \cite{Bai2021} and medical image segmentation \cite{zamzmi2021trilateral}. Initial testing of this model shows a favorable performance profile when deployed against two state-of-the-art edge devices, the NVIDIA Jetson Nano and Xavier, as shown in Table \ref{processing-speed-table}. For the Nano, we tested at two power levels. For both devices, we tested at varying clock speeds with and without TensorRT, a deep learning inference optimization library. Each of these devices were then tested for steady-state frames-per-second at varying input image sizes. As can be seen across all examples, the frame rate decreases rapidly as the input size doubles. Even with the largest image size we tested, however, the Xavier operating at the max clock speed with TensorRT can still operate at almost 2 frames per second, a suitable frame rate given the altitude and field of view of many survey missions. Detailed profiling results at more granular resolution step sizes, taken across all settings of power and TensorRT optimizations, are provided within the Github repository for this project.

\subsection{Active Learning Annotation Workflow}
\label{activelearning}

Our initial investigations started with a completely manual annotation process, drawing polygons to delineate regions. However, due to the intricate and jagged regions typical of snow and ice, accurately annotating boundaries in this manner was very time consuming. We explored available annotation assistance algorithms, such as GrabCut, CVAT Intelligent Scissors, and iterative training guided by extreme-point selection~\cite{reviving2021}. While these algorithms can be efficient for segmentation annotation in more common domains, we found them to be inefficient on our data due to their extensively segmented and complex morphology. Zhang et. al observed similar challenges and chose to use Photoshop for dense annotation~\cite{ICENET2020}.

\begin{figure*}[t]
  \centering
  \includegraphics[width=\textwidth]{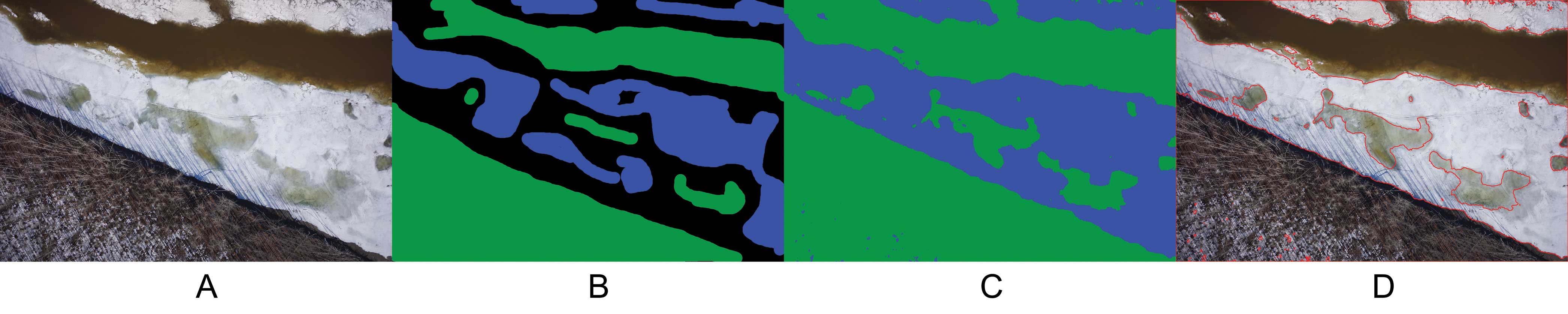}
  \caption{(A) Example raw image from KUAF River Ice Dataset. (B) Sparse annotation associated with the image. Region interiors are "painted" with the appropriate labels without addressing the complex boundaries.  Blue corresponds to frozen water (both snow and ice) and green corresponds to the background (land and liquid water). Black indicates space where no label was given. (C) After some initial training solely on sparsely annotated examples, the model can successfully interpolate the remainder of the labels (black regions disappear), which after visual validation, was accepted as ground truth. (D) The region boundaries from C shown as red contours on top of original image. Figure is best viewed in color.}
  \label{fig:sparse_ann}

\end{figure*}

Given that ice and snow are relatively distinct from most backgrounds, both in terms of color and texture, we postulated that even with limited training data, a segmentation model should generalize well when applied to visually similar examples. This inspired us to develop and deploy a progressive label correction (active learning) workflow. We start by very coarsely and sparsely manually "painting" region labels using the open-source image editor GIMP. We paint the labels in a separate layer with the class specified by the choice of color. We do not attempt to extend these annotations fully to the complex region boundaries. Figure~\ref{fig:sparse_ann}B shows an example of this sparse annotation with frozen water (both snow and ice) marked in purple, regions that are not frozen water marked in green, and unannotated regions shown in black. We build up sparse annotations in this manner for 5 images and then train the segmentation model (unannotated regions do not contribute to training loss). We then deploy that model back on the sparsely annotated images as well as completely new images. In this way, we use the model to make suggestions for the full dense ground truth for these images (Figure~\ref{fig:sparse_ann}C). We do not expect or need this model output to be perfect, we simply need it to be correct over a sufficient fraction of the images such that a person checking and correcting mistakes is quicker than annotating from scratch.

\begin{table}[t]
\centering
\begin{tabular}{|c|c|c|c|}
    \hline
    \textbf{Input Size} & \textbf{7.5W Nano} & \textbf{10W Nano} & \textbf{Xavier w/TRT} \\
    \hline
    256x256&11.64&12.12&111.8 \\
    512x512&4.58&4.74&36.68 \\
    1024x1024&0.83&0.88&10.14 \\ 
    2048x2048&0.14&0.19&1.99 \\
    \hline
\end{tabular}
\caption{Frame rates (Hz) at which the payload runs our ice segmentation model for different input sizes and devices.}
\label{processing-speed-table}
\end{table}

Our tools allow us to quickly inspect the quality of the model-suggested annotations for each image, and we sort these results into 1) \textit{ground-truth-ready}, 2) \textit{minor-corrections-needed}, and 3) \textit{hard-negative}. We were able to accept ground-truth-ready images as dense ground truth 56\% of the time. We found 19\% of examples required only minor corrections, and these are loaded back into GIMP to quickly clean up the residual labeling errors. Identified hard negatives are chosen for sparse annotation, if not already done so, or complete manual dense annotation.

As we iterated through this annotation process and trained updated versions of the model with more validated ground truth, the model generalized better on the yet-unannotated examples, we spent less time correction residual errors, and ground-truth generation became more efficient. With this annotation workflow, the annotator efficiently focuses effort on regions and examples where the current model is ineffective and does not waste effort on examples where the model is already very accurate. 

\begin{figure}[!t]
  \centering
  \includegraphics[width=\linewidth]{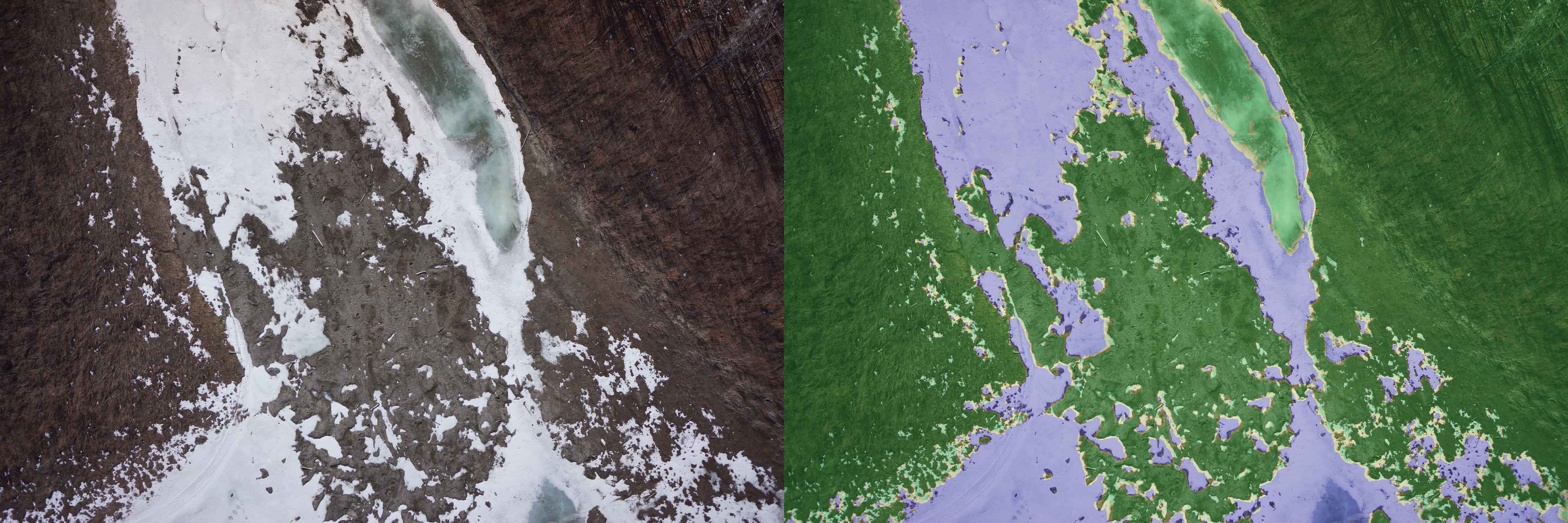}
  \caption{Results of a model trained on binary ice/no-ice ground truth data. (Left) Raw image from KUAF River Ice Dataset. (Right) A segmentation output showing the ice (blue) and non-ice (green) regions as determined by the model. Figure is best viewed in color.}
  \label{fig:ice_seg}
\end{figure}

Sparse annotations took 2.02 minutes per image on average plus 1.03 minutes to clean up the model's interpolated labels. Additionally, half of the final ground-truth image annotations were drawn directly, after quick visual validation, from the segmentation model's output without any manual annotation or correction whatsoever. This is in contrast to and a considerable savings from the average of 8.52 minutes per image required to achieve similar-quality results by densely annotating from scratch. After all ground truth was assembled, the model was re-trained from scratch.

We trained the BiSeNetV2 model on our ice segmentation dataset for 100 epochs on random patches of 500×500 pixels, cropped out of a 2x downsampled version of the original 8000×6000 pixel imagery. Random flips, rotations, and input normalization were used as augmentation. We used the ADAM optimizer with learning rate 0.02, warm up for 500 iterations, and a step of 1e-1 after epochs 40 and 70. The batch size used was 18. For this effort, we used only a binary ice/no-ice class training. The results of this training are shown in Figure \ref{fig:ice_seg}. 

\section{Payload Flight Validation}
\label{flightval}

\begin{figure*}[!t]
  \centering
  \includegraphics[width=\linewidth]{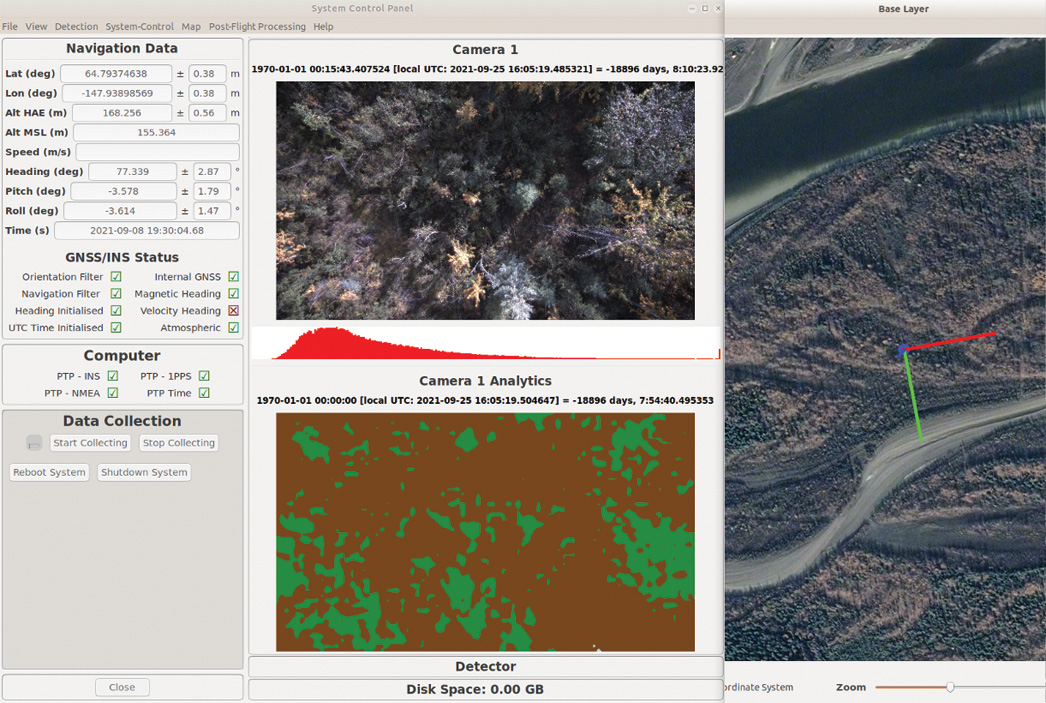}
  \caption{ADAPT payload deployment over the Tanana river in Alaska (September 2021). Screen capture from base-station laptop of the live view of ADAPT payload operation, while 665 meters away, showing a reduced-resolution view of the live imagery and live segmentation model outputs, the image histogram, payload pose shown on a geospatial base layer, and various diagnostics. Figure best viewed in color. Base layer imagery \textcopyright\ 2021 Maxar Technologies and CNES/Airbus.}
  \label{fig:payloadAlaska}
\end{figure*}

Test flights were conducted in Fairbanks, Alaska in September 2021 to validate system capabilities. Since these were warm-weather flights without snow or ice, we deployed a proxy tree-segmentation model trained using data synthetically generated using Microsoft AirSim. This model has an identical network architecture and number of parameters as our ice-segmentation DNN, so it is a realistic stress test for real-time analytics generation and broadcast. We flew a 16.1 megapixel FLIR Blackfly (5320 × 3032 px) with an 8-mm f/1.8 lens, mounted nadir-looking, capturing 4 frames per second. The flight controller was programmed to autonomously cover a lawnmower flight pattern with cross-track overlap. Flights were conducted along the Tanana river at two locations: One with the unmanned system flying at approximately 30 m altitude above ground level (AGL) with a 1 cm ground sample distance (GSD) approximately covering a 600 m x 1,100 m area extending to $\sim$1,251 m from the base station, and the other at approximately 60 m AGL with a 2 cm GSD approximately covering a 1,200 m x 1,200 m area. This coverage was achieved after 138 minutes of flying with four changes of the drone's dual 22,000 mAh batteries (average flight time of 27.6 minutes).

An additional flight was conducted with the camera oriented $\sim$45\textdegree\ down from forward, and the drone was programmed to fly up along the river bank and back. Pairs of views covering each section of the bank, drawn from the forward and return flight path, generate sets of wide-baseline stereo pairs, optimal data for SfM topological reconstruction of the river bank. This provides data with which to study the mission requirements for monitoring coastline erosion.

Figure~\ref{fig:payloadAlaska} shows a live screen capture of the ADAPT graphical user interface (GUI) running on our base-station laptop communicating via WiFi with the payload at a distance of 665 meters. The GUI receives reduced-resolution and highly compressed views of the high-resolution raw imagery and associated segmentation outputs generated on-board the payload. We can also monitor payload operation via a variety of diagnostics. A remote view of the image histogram helps to ensure that the camera is configured with a proper exposure setting. The GUI also provides a separate panel to inspect image sharpness, facilitating on-the-ground camera focusing and in-flight validation. This feedback was particularly useful in identifying the onset of motion-induced blur at exposures longer than 2 milliseconds. Reducing the maximum exposure time to 500 $\mu$s prevented further issues, even during 10 m/s forward flight.

During testing, we maintained WiFi connectivity out to 1 km, only losing momentary connection when wireless line-of-site to the sUAS was blocked by treeline (safety observers stationed elsewhere maintained visibility throughout). We verified that the payload remained operational during communications blackouts. We anticipate that extended beyond-line-of-site operation would require alternative wireless communication systems, which tend to have reduced bandwidth. Therefore, we also verified that segmentation results could be compressed via boundary vectorization down to 20 KB per image with a negligible loss in accuracy, retaining full utility even with reduced bandwidth. Over 800 GB of high resolution imagery was collected along with time-synchronized INS data (at approximately 100Hz). Various states of the river and riverbank boundary were observed along with various forms of vegetation and plant life.  This data will be made publicly available.

\section{Conclusion and Future Direction}
In this paper, we presented a novel open-source system for sUAS mission planning and development. We validated that the payload can be provisioned with readily off-the-shelf components, assembled, loaded with a state of the art deep learning model for semantic segmentation, and flown in order to provide real time in-situ intelligence with minimal upfront non-recurring engineering cost. We make all hardware plans, software, and network training code needed to replicate these results free and open-source to the community with the aim of reducing the time and capital necessary to provision an sUAS mission in the future. We hope and will be advocating for the community to adopt this baseline result and build more features and payload configurations into the ADAPT system. For future work, we plan further feature development of the open-source system in support of flight odometery and multi-sensor fusion. Additionally, we plan to continue development of the river ice segmentation dataset and model in order to solve the pressing issue of river ice breakup detection and flood prediction during wet seasons.  

\section{Ethical statement}
As discussed in this paper, there are many positive impacts to be realized from an open-source, adaptable, AI-enabled sUAS payload. It must be considered that in the making of drone and AI technology more accessible to those trying to improve the human condition, we also enable those who would seek to thwart it. While AI-enabled mobile platforms can be used for a wide range of good in HADR, they can also be used to more efficiently poach endangered wildlife, counteract law enforcement and response personnel, carry out acts of violence, or collect data on marginalized populations for the purposes of disenfranchisement or exploitation. It is the duty of those contributing to the democratization of AI and sUAS capabilities to be aware of potential blind spots in our society's technological understanding of these elements and advocate for policies, rules, and regulations which protect those most vulnerable to the potential for abuse. 

\section{Acknowledgements}
This paper was prepared by Kitware, Inc. using Federal funds under award NA20OAR0210083 from the National Oceanic and Atmospheric Administration, U.S. Department of Commerce. The statements, findings, conclusions, and recommendations are those of the author(s) and do not necessarily reflect the views of the National Oceanic and Atmospheric Administration or the U.S. Department of Commerce.

\bibliography{aaai22.bib}

\end{document}